\documentclass[10pt]{elsart}

\usepackage{amsmath}
\usepackage{amssymb}
\usepackage{cite}
\usepackage{array}
\usepackage{enumerate}
\usepackage{mathrsfs}
\journal{AMC}

\begin{document}

\begin{frontmatter}



\title{\textbf{A Weight-coded Evolutionary Algorithm for the Multidimensional Knapsack Problem}}


\author[wayne]{Quan Yuan\corauthref{cor}},~\ead{quanyuan@wayne.edu}
\author[uwec]{Zhixin Yang}
\corauth[cor]{Corresponding authors.} \address[wayne]{Department of
Mathematics, Wayne State University,\\ Detroit, MI 48202, USA }
\address[uwec]{Department of Mathematics, University of Wisconsin-Eau Claire,\\ Eau Claire, WI 54702, USA}

\begin{abstract}
A revised weight-coded evolutionary algorithm (RWCEA) is proposed
for solving multidimensional knapsack problems. This RWCEA uses a
new decoding method and incorporates a heuristic method in
initialization. Computational results show that the RWCEA performs
better than a weight-coded evolutionary algorithm proposed by Raidl
(1999) and to some existing benchmarks, it can yield better results
than the ones reported in the OR-library. \end{abstract}

 \end{frontmatter}

\section{Introduction}
The multidimensional knapsack problem (MKP) can be stated as:%
\begin{subequations}\label{eq1}%
\begin{align}%
\max & \,f(x)=\sum^n_{j=1}p_jx_j, &\label{eq1-1}\\
\text{s.t.} & \sum^n_{j=1}r_{ij}x_j \le b_i, &
i=1,\ldots,m,\label{eq1-2}\\
& x_j\in \{0,\,1\}, &
j=1,\ldots,n.\label{eq1-3}
\end{align}%
\end{subequations}%
Each of the $m$ constraints described in \eqref{eq1-2} is called a
{\em knapsack} constraint. A set of $n$ items with profits $p_j>0$
and $m$ resources with $b_i>0$ are given. Each item $j$ consumes an
amount $r_{ij}\ge0$ from each resource $i$. The 0-1 decision
variables $x_j$ indicate which items are selected. A {\em
well-stated} MKP also assumes that $r_{ij}\le
b_i<\sum^n_{j=1}r_{ij}$ and $p_j>0$ for all $i\in I=\{1,\ldots,m\}$,
$j\in J=\{1,\ldots,n\}$, since any violation of these conditions
will result in some constraints being eliminated or some $x_j$'s
being fixed.

The MKP degenerates to the {\em knapsack problem} when $m=1$ in Eq.
\eqref{eq1-2}. It is well known that the knapsack problem is not a
strong $\mathcal{NP}$-hard problem and solvable in pseudo-polynomial
time. However, the situation is different to the general case of $m
> 1$. Garey and Johnson (1979)\cite{garey1979} proved that it is
strongly $\mathcal{NP}$-hard and exact techniques are in practice
only applicable to instances of small to moderate size.

A real-world application example of MKP is selecting projects to
fund. Assume there are $n$ different projects and we need to select
some projects and fund them for $m$ years. Each project provides a
profit and each of them has a budget determined for each year. Our
objective is to maximize the total profit and not exceed yearly
budgets. This problem can be formulated as Eq. \eqref{eq1}. What is
more, many practical problems such as the capital budgeting
problem\cite{manne195784}, allocating processors and databases in a
distributed computer system\cite{gavish1982215}, project selection
and cargo loading \cite{shih1979369}, and cutting stock
problems\cite{gilmore19661045} can be formulated as an MKP. The MKP
is also a subproblem of many general integer programs.

Given the theoretical and practical importance of the MKP, a large
number of papers have devoted to the problem. It is not the place
here to recall all of these papers. We refer to the papers of Chu
and Beasley (1998)\cite{chu199863}, Fr\'{e}ville
(2004)\cite{freville20041} and the monograph of Kellerer
(2004)\cite{kellerer2004} for excellent overviews of theoretical
analysis, exact methods, and heuristics of the MKP. Recently, some
new algorithms for the MKP have been proposed such as some variants
of the genetic algorithm\cite{li2006696}, the ant colony
algorithm\cite{kong20082672}, the scatter search
method\cite{hanafi2008143}, and some new
heuristics\cite{boyer2008inpress,fleszar20091602,puchinger2005,zou20111556}.
Some studies on analysis of the
MKP\cite{raidl2005441,freville2005195} and generalizations of the
MKP\cite{gorski2011,captivo20031865,hifi20051311} have also been put
forward.

An {\em Evolutionary algorithm} (EA) is a generic population-based
metaheuristic optimization algorithm. Candidate solutions to the
optimization problem play the role of individuals (parents) in a
population. Some mechanisms inspired by biological evolution:
selection, crossover and mutation are used. The fitness function
determines the environment within which the solutions ``survive''.
Then new groups of the population (children) are generated after the
repeated application of the above operators. EAs have found
application in computational science, engineering, economics,
chemistry, and many other fields (See
\cite{yuan2010640,yuan2008924,yuan201036,yuan2008257,yuan201311408,zhu20131}).

In the last two decades EAs were studied for solving the MKP.
Although the early works do not successfully show that {\em genetic
algorithms} (GAs) were an effective tool for the MKP, the first
successful GA's implementation was proposed by Chu and Beasley
(1998)\cite{chu199863}. Extended numerical comparisons with
\texttt{CPLEX} (version 4.0) and other heuristic methods showed that
Chu and Beasley's GA has a robust behavior and can obtain
high-quality solutions within a reasonable amount of computational
time. Raidl and Gottlieb (2005)\cite{raidl2005441} introduced and
compared six different EAs for the MKP, and performed static and
dynamic analyses explaining the success or failure of these
algorithms, respectively. They concluded that an EA based on direct
representation, combined with local heuristic improvement (referred
to as DIH in \cite{raidl2005441}, i.e., GA of Chu and Beasley
(1998)\cite{chu199863} with slight revision), can achieve better
performance than other EAs mentioned in \cite{raidl2005441} from
empirical analysis.

The best success for solving the MKP, as far as we known, has
been obtained with tabu-search algorithms embedding effective
preprocessing\cite{vasquez2001328,vasquez200570}. Recently,
impressive results have also been obtained by an implicit
enumeration\cite{vimont2008165}, a convergent
algorithm\cite{hanafi2009}, and an exact method based on a
multi-level search strategy\cite{boussier201097}. Compared with
EAs, the methods mentioned above can yield better results when
excellent solutions are required. But they are more complicated
to implement or their computation takes extremely long time.
Since EAs are simple to implement and their computation time are
easy to control, they are good alternatives if the quality
requirement of solutions of the MKP is not very strict.

In this paper, we will consider a variant of EA to solve the MKP.
This EA will use a special encoding technique which is called {\em
weight-coding} (or {\em weight-biasing}). We will revise a
weight-coded EA (WCEA) proposed by Raidl (1999)\cite{raidl1999596}
and propose a revised weight-coded EA (RWCEA). The numerical
experiments of some benchmarks will show that the RWCEA performs
better than the WCEA. Moreover, this RWCEA can compete with DIH in
some benchmarks.

\section{An introduction to the weight-coding and its
application to the MKP} When combinatorial optimization problems
are solved by an EA, the coding of candidate solutions is a
preliminary step. Direct coding such as the {\em binary coding}
is an intuitive method. The main drawback of this coding lies in
that many infeasible solutions may be generated by EA's
operators. To avoid that, the basic idea of the weight-coding is
to represent a candidate solution by a vector of real-valued
weights $w_j\,(j=1,\ldots,n)$. The {\em phenotype} that a weight
vector represents is obtained by a two-step process.
\begin{enumerate}[Step (a):] \item ({\em biasing}) The original
problem $P$ is temporarily modified to $P'$ by biasing problem
parameters of $P$ according to the weights $w_j$; \item ({\em
decoding heuristic}) A problem-specific decoding heuristic is
used to generate a solution to $P'$. This solution is
interpreted and evaluated for the original (unbiased) problem
$P$. \end{enumerate}

The weight-coding is an interesting approach because it can
eliminate the necessity of an explicit repair algorithm, a
penalization of infeasible solutions, or special crossover and
mutation operators. It has already been successfully used for a
variety of problems such as an optimum communications spanning
tree problem\cite{palmer1994379}, 
problem\cite{capp1998327}, the traveling salesman
problem\cite{julstrom1998313}, and the multiple container
packing problem\cite{raidl1999}.

To the best of the authors' knowledge, the work of Raidl
(1999)\cite{raidl1999596} is the first to use weight-coded EA
(WCEA) to deal with the MKP. In that paper, some variants of
WCEAs were proposed and compared. And Raidl finally suggested
one of them and compared the WCEA with other EAs in
\cite{raidl2005441}. In this WCEA, $w_j\,(j=1,\ldots,n)$ is set
to be the weight vector representing a candidate solution.
Weight $w_j$ is associated with item $j$ of the MKP.
Corresponding to Step (a), the original MKP is biased by
multiplying of profits in \eqref{eq1-1} with {\em log-normally}
distributed weights: \begin{equation}\label{eq4}
p'_j=p_jw_j=p_j(1+\gamma)^{\mathcal {N}(0,1)},\;j=1,\ldots,n
\end{equation} where $\mathcal{N}(0,1)$ denotes a normally
distributed random number with mean $0$ and standard deviation
$1$, and $\gamma>0$ is a strategy parameter that controls the
average intensity of biasing. Raidl (1999)\cite{raidl1999596}
suggested that $\gamma=0.05$. Since the resource consumption
values $r_{ij}$ and resource limits $b_i$ are not modified, all
feasible solutions of the biased MKP are feasible to
\eqref{eq1}.

Corresponding to Step (b), the decoding heuristic which Raidl
(1999)\cite{raidl1999596} suggested is making use of the {\em
surrogate relaxation} (See \cite{hanafi1998659,glover1975434}).
The $m$ resource constraints \eqref{eq1-2} are aggregated into a
single constraint using surrogate multipliers $a_i$, $i =
1,\ldots,m$: \begin{equation}
\sum^n_{j=1}\left(\sum^m_{i=1}a_ir_{ij}\right)x_j\le
\sum^m_{i=1}a_ib_i \end{equation} where $a_i$ are obtained by
solving the linear programming (LP) of the relaxed MKP, in which
the variables $x_j$ may get real values from $[0,1]$. The values
of the dual variables are then used as surrogate multipliers,
i.e. $a_i$ is set to the shadow price of the $i$-th constraint
in the LP-relaxed MKP. {\em Pseudo-utility ratios} are defined
as: \begin{equation}\label{eq6}
u_j=\frac{p_j'}{\sum^m_{i=1}a_ir_{ij}}. \end{equation} A higher
pseudo-utility ratio heuristically indicates that an item is
more efficient. After the items are sorted by decreasing order
of $u_j$, the {\em first-fit strategy} used as decoder in the
permutation representation is applied. All items are checked one
by one and each item's variable $x_j$ is set to $1$ if no
resource constraint is violated, otherwise, $x_j$ is set to $0$.
The computational effort of the decoder is $O(n\cdot \log n)$
for sorting the $u_j$ plus $O(n\cdot m)$ for the first-fit
strategy, yielding $O(n\cdot(m+\log n))$ in total.

Raidl's WCEA can be described as follows (we will explain the
details of Steps 6, 7, and 8 afterward):

\textbf{Algorithm of Raidl's WCEA} \begin{enumerate}[Step 1:]
\item set $t:=0$; \item initialize $pop(t)=\{S_1,\ldots,S_N\}$,
$S_i=(w_1,\ldots,w_n)$ where $w_j$ is a random value following
log-normally distribution as \eqref{eq4}; \item evaluate
$pop(t):\,\{f(S_1),\ldots,f(S_N)\}$;\\ for each $S_i$
\begin{enumerate}[{3-}1:] \item bias original MKP; \item use
decoding heuristic as in \cite{raidl1999596} (described above)
to get phenotype $\mathscr{P}(S_i)\in \{0,\,1\}^n$; \item
substitute $\mathscr{P}(S_i)$ into \eqref{eq1-1} to obtain
$f(S_i)$; \end{enumerate} \item find $S^*\in pop(t)$ $s.t.$
$f(S^*)\ge f(S)$, $\forall$ $S\in pop(t)$;
$t<t_{\max}$ \textbf{do} \item select $\{p_1,\,p_2\}$ from
$pop(t)$; \item crossover $p_1$ and $p_2$ to generate a child
$C$; \item mutate $C$; \item evaluate $C$ as Step 3, get
$\mathscr{P}(C)$ and $f(C)$; \item \textbf{if}
$\mathscr{P}(C)\equiv$ any $\mathscr{P}(S_i)$ \textbf{then}
(that means $C$ is a duplicate of a member of the population)
\item \ \ discard $C$ and goto Step 6;\\ \textbf{end if} \item
find $S'\in pop(t)$ $s.t.$ $f(S')\le f(S)$ $\forall S\in pop(t)$
and replace $S'\leftarrow C$; ({\em steady-state replacement},
i.e., the worst individual of population is replaced.) \item
\textbf{if} $f(C)>f(S^*)$ \textbf{then} \item \ \ $S^*\leftarrow
C$; (update best solution $S^*$ found)\\ \textbf{end if} \item
$t\leftarrow t+1$;\\ \textbf{end while} \item return $S^*$,
$f(S^*)$. \end{enumerate}

In Step 6, a {\em binary tournament selection} is used. That is,
two pools of individuals, which consist of $2$ individuals drawn
from the population randomly, are formed respectively at first.
Then two individuals with the best fitness, each taken from one
of the two tournament pools, are chosen to be parents.

In Step 7, Raidl (1999)\cite{raidl1999596} suggested a {\em
uniform crossover} instead of one- or two-point crossover. In
the uniform crossover two parents have one child. Each
$w_j(j=1,\ldots,n)$ in the child is chosen randomly by copying
the corresponding weight from one or the other parent.

Once a child has been generated through the crossover, a {\em
mutation} step in Step 8 is performed. Each $w_j$ of the child
is reset to a new random value observing log-normal distribution
with a small probability ($3/n$ per weight as in
\cite{raidl1999596} or one random position in
\cite{raidl2005441}).

In numerical experiments, the $N$ in Step 2 is taken as $100$
and $t_{\max}$ in Step 5 is taken $10^6$. Raidl and Gottlieb
(2005)\cite{raidl2005441} compared this WCEA with other five EAs
for the MKP. From empirical analysis, this WCEA outperformed all
of them except DIH (The meaning of DIH is given in Section 1) on
average.

\section{Our revised WCEA for the MKP} \subsection{Motivation}
The core of Raidl's WCEA is the surrogate relaxation based
heuristic in decoding. In our points of view, this heuristic has
two drawbacks. First, the dual variables of an LP-relaxed MKP
used in heuristic decoding step are just good approximations of
optimal surrogate multipliers and it may mislead the
search\cite{vasquez2001328}. 
LP-relaxed MKP used in heuristic decoding step are just 
approximations of optimal surrogate multipliers. And deriving
optimal surrogate multipliers is a difficult task in
practice\cite{gavish198578}. Secondly, the heuristic decoding
might mislead the search if the optimal solution is not very
similar to the solution generated by applying the greedy
heuristic\cite{rothlauf2003381}.

In order to avoid using surrogate multipliers, we set
$w_j\,(j=1,\ldots,n)$ to let every $w_j$ observe uniform
distribution on $[0,\,p_{\max}/p_j]$, where
$p_{\max}=\max\{p_j:j=1,\,\ldots,\,n\}$. The profits of the
original MKP are biased by multiplying weights:
\begin{equation}\label{eq7} p'_j=p_jw_j,\;j=1,\ldots,n.
\end{equation} as mentioned in Section II, all feasible
solutions of this biased MKP are feasible to \eqref{eq1}. In
decoding heuristic, we also use first-fit strategy, i.e., the
items are sorted by decreasing order of $p'_j$ (not by
pseudo-utility ratio in \eqref{eq6}) and traversed. Each item's
variable $x_j$ is set to $1$ if no resource constraint is
violated. The computational effort of the decoder is also $O(n
\cdot (m + \log n))$ in total.

This form of $w_j$ is similar to the idea of {\em Random-key
Representation}\cite{hinterding19991286}. Surrogate multipliers
can be avoided but the efficiency of the EA will be
reduced\cite{raidl2005441}. To overcome this disadvantage, our
thought is to obtain a ``good'' initial population. In the
following we first introduce an idea proposed by Vasquez and
Hao\cite{vasquez2001328} and then propose our method.

It is well known that only relaxing the integrality constraints
in an MKP may not be sufficient because its optimal solution may
be far away from the optimal binary solution. However, Vasquez
and Hao in \cite{vasquez2001328} observed when the integrality
constraints was replaced by a {\em hyperplane constraint}
$\sum^n_{j=1}x_j=k\in \mathbb{N}$, the corresponding linear
programming solution may often be close to the optimal binary
solution. For example in \cite{vasquez2001328}, in \eqref{eq1}
we let $n=5$, $m=1$, $\boldsymbol{p}=\{12,\,12,\,9,\,8,\,8\}$,
$\boldsymbol{r}=\{11,\,12,\,10,\,10,\,10\}$, $b=30$. The relax
linear programming problem leads to the fractional optimal
solution $x^{LP}=\{1,\,1,\,0.7,\,0,\,0\}$ while the optimal
binary solution is $x=\{0,\,0,\,1,\,1,\,1\}$. If we replace the
integrality constraints by $\sum^n_{j=1}x_j=3$, this linear
programming problem leads to the optimal binary solution.

In the above example, if we take
$\boldsymbol{w}=\{0,\,0,\,1,\,1,\,1\}$ and substitute it to
\eqref{eq7}, the optimal binary solution can be obtained by
first-fit heuristic mentioned above. Moreover, if we do not
restrict $k$ as an integer, we may also obtain some
corresponding linear programming solutions from which some good
binary solutions may be obtained by first-fit heuristic. We use
these linear programming solutions as a ``good'' initial
population. So the disadvantage of Random-key Representation may
be overcome. The experimental results presented later have
confirmed this hypothesis. Naturally, the hypothesis does not
exclude the possibility that there exists a certain MKP whose
optimal binary solution cannot be obtained from linear
programming solutions.

Inspired by this idea, initialization is guided by the LP
relaxation with a hyperplane constraint. To begin with, we use
some simple heuristic (such as a greedy algorithm) to obtain a
0-1 lower bound $z$. Next, the two following problems:
\begin{align*} & \;k_{\max}=\max\sum^n_{j=1}x_j, &\\ \text{s.t.}
& \sum^n_{j=1}r_{ij}x_j \le b_i, & i=1,\ldots,m,\\ &
\sum^n_{j=1}p_jx_j\ge z+1 &\\ & x_j\in [0,1] & j=1,\ldots,n
\end{align*} and \begin{align*} & \;k_{\min}=\min
\sum^n_{j=1}x_j, &\\ \text{s.t.} & \sum^n_{j=1}r_{ij}x_j \le
b_i, & i=1,\ldots,m,\\ & \sum^n_{j=1}p_jx_j\ge z+1 &\\ & x_j\in
[0,1] & j=1,\ldots,n \end{align*} are solved to obtain
$k_{\max}$ and $k_{\min}$.

Then, $N$ linear programming problems
\begin{equation}\label{n-lp}\begin{aligned} \max & \sum^n_{j=1}p_jx_j, &\\
\text{s.t.} & \sum^n_{j=1}r_{ij}x_j \le b_i, & i=1,\ldots,m,\\ &
\sum^n_{j=1}x_j=k' &\\ & x_j\in [0,1] & j=1,\ldots,n
\end{aligned}\end{equation} are solved where $k'$ is a real number generated
randomly from $[k_{\min},k_{\max}]$ in each computation. So the $N$
linear programming solutions are generated as the initial
population.

\subsection{Implementation}
The scheme of the RWCEA is as follows:

\textbf{Algorithm of the RWCEA}
\begin{enumerate}[Step 1:]
\item set $t:=0$;
\item initialize $pop(t)=\{S_1,\ldots,S_N\}$ by solving $N$ linear programming problems of \eqref{n-lp},
$S_i=(w_1,\ldots,w_n)$ where $w_j$ is a random value following
uniform distribution on $[0,p_{\max}/p_j$, where
$p_{\max}=\max\{p_j:\,j=1,\ldots,n\}$;
\item evaluate $pop(t):\,\{f(S_1),\ldots,f(S_N)\}$;\\
for each $S_i$
\begin{enumerate}[{3-}1:]
\item bias original MKP;
\item use decoding heuristic as in \cite{raidl1999596} (described in Section 2) to
get phenotype $\mathscr{P}(S_i)\in \{0,\,1\}^n$;
\item substitute
$\mathscr{P}(S_i)$ into \eqref{eq1-1} to obtain $f(S_i)$;
\end{enumerate}
\item find $S^*\in pop(t)$ $s.t.$
$f(S^*)\ge f(S)$, $\forall$ $S\in pop(t)$;
$t<t_{\max}$ \textbf{do}
\item select $\{p_1,\,p_2\}$ from $pop(t)$;
\item crossover $p_1$ and $p_2$ to generate a child $C$;
\item mutate $C$: one random $w_j$ of the child is reset to a new
random value observing uniform distribution on $[0,\,p_{\max}/p_j]$;
\item evaluate $C$ as Step 3, get $\mathscr{P}(C)$ and
$f(C)$; \item \textbf{if} $\mathscr{P}(C)\equiv$ any
$\mathscr{P}(S_i)$ \textbf{then} (that means $C$ is a duplicate of a
member of the population)
\item \ \ discard $C$ and goto Step 6;\\ \textbf{end if} \item
find $S'\in pop(t)$ $s.t.$ $f(S')\le f(S)$ $\forall S\in pop(t)$ and
replace $S'\leftarrow C$; ({\em steady-state replacement}, i.e., the
worst individual of population is replaced.) \item \textbf{if}
$f(C)>f(S^*)$ \textbf{then} \item \ \ $S^*\leftarrow C$; (update
best solution $S^*$ found)\\ \textbf{end if} \item $t\leftarrow
t+1$;\\ \textbf{end while} \item return $S^*$, $f(S^*)$.
\end{enumerate}

The scheme of the RWCEA is similar to Raidl's WCEA. And we take the
same values of $N$ and $t_{\max}$ as the WCEA. The differences
between the two algorithms lie in the following aspects:
\begin{enumerate}
\item The initial
population in Raidl's WCEA is generated randomly, while in the
RWCEA, $N$ linear programming problems should be solved;
\item Each $w_j$ in Raidl's WCEA observes
log-normal distribution, while in RWCEA it observes a uniform
distribution on $[0,\,p_{\max}/p_j]$, where
$p_{\max}=\max\{p_j:j=1,\,\ldots,\,n\}$;
\item Raidl's WCEA sorts
items by pseudo-utility ratios in heuristic decoding step while the
RWCEA sorts items by biased profits directly;
\item In
the mutation step, one random $w_j$ of the child is reset to a new
random value observing uniform distribution on $[0,\,p_{\max}/p_j]$
instead of log-normal distribution in the RWCEA. \end{enumerate}

In summary, we revised Raidl's WCEA by avoiding using surrogate
multipliers and using ``good'' initial population. We think this
RWCEA can yield better result than WCEA in some instances of MKP.
The performance of RWCEA is shown in the next section.

\section{Experimental comparison}
As in \cite{raidl2005441}, two test suites of MKP's benchmark
instances for experimental comparison are used in this paper. The
first one, referred to as CB-suite in this paper, is introduced by
Chu and Beasley (1998)\cite{chu199863} and is available in the
OR-Library\footnote{http://people.brunel.ac.uk/$\sim$mastjjb/jeb/info.html}.
This test suite contains $270$ instances for each 10 ones are
combination of $m\in\{5, 10, 30\}$ constraints, $n\in \{100, 250,
500\}$ items, and tightness ratio $\alpha \in \{0.25, 0.5, 0.75\}$.
Each problem has been generated randomly such that $b_i
=\alpha\cdot\sum^n_{j=1}r_{ij}$ for all $i = 1,\ldots,m$. Chu and
Beasley used their GA (i.e., DIH) to solve these instances and
reported their results in the OR-library. The second MKP's benchmark
suite\footnote{ This suite can be downloaded from
http://hces.bus.olemiss.edu/tools.html} used in \cite{raidl2005441}
was first referenced by \cite{vasquez2001328} and originally
provided by Glover and Kochenberger. These instances, called GK01 to
GK11, range from $100$ to $2500$ items and from $15$ to $100$
constraints. We call this suite GK-suite in this paper.

\begin{table} \renewcommand{\arraystretch}{.9}
\begin{tabular}[c]{cccccccccc}
\hline
\multicolumn{3}{c} {instance} & \multicolumn{7}{c} {gap[\%](and standard deviation)}\\
\hline
name & $m$ & $n$ & PE & OR & RK & DI & WB & DIH & RWCEA\\
\hline
 CB1 & 5  & 100 & 0.425 & 0.745 & 0.425 & 0.425 & 0.425  & 0.425& 0.425\\
 &&& (0.000) & (0.210) & (0.000) & (0.000) & (0.000) &(0.000) & ({\bf0.000})\\
 \hline
 CB2 & 5  & 250 & 0.120 & 1.321 & 0.115 & 0.150 &0.106  & 0.106& 0.112\\
 &&& (0.012) & (0.346) & (0.009) & (0.019) & (0.007) & (0.006) &({\em 0.007})\\
 \hline
 CB3 & 5  & 500 & 0.081 & 2.382 & 0.065 & 0.121 &0.042  & 0.038& 0.036\\
&&& (0.016) & (0.657)&(0.010)&(0.020) & (0.008)&(0.003)&({\em 0.004})\\
\hline
 CB4 & 10  & 100 & 0.762& 1.013&0.762&0.770& 0.761  & 0.762& 0.762\\
&&& (0.001)&(0.163)&(0.003)&(0.013)&(0.000)&(0.003)&({\bf0.003})\\
\hline
 CB5 & 10  & 250 &0.295&1.498&0.277&0.324& 0.249  & 0.261& 0.271\\
 &&&(0.033) &(0.225)&(0.021)&(0.043)&(0.017)&(0.008)&({\em0.014})\\
 \hline
 CB6 & 10  & 500 & 0.225&2.815&0.200&0.263&0.131  & 0.112& 0.108\\
 &&&(0.040) &(0.462)&(0.029)&(0.040)&(0.014)&(0.007)&({\bf0.002})\\
 \hline
 CB7 & 30  & 100 &1.372&1.800&1.338&1.401& 1.319  & 1.336& 1.276\\
 &&&(0.134) &(0.182)&(0.123)&(0.073)&(0.093)&(0.091)&({\bf0.077})\\
 \hline
 CB8 & 30  & 250 &0.608&2.076&0.611&0.599& 0.535  & 0.519& 0.525\\
 &&&(0.048) &(0.346)&(0.072)&(0.059)&(0.031)&(0.013)&({\bf0.002})\\
 \hline
 CB9 & 30  & 500 &0.429&3.267&0.376&0.463& 0.306  & 0.288& 0.296\\
 &&&(0.058) &(0.442)&(0.037)&(0.056)&(0.024)&(0.012)&({\bf0.012})\\
 \hline
 GK01 & 15  & 100 &0.377&0.683&0.384&0.336& 0.308  & 0.270& 0.325\\
&&& (0.068)&(0.098)&(0.080)&(0.074)&(0.077)&(0.028)&({\em0.077})\\
\hline
 GK02 & 25  & 100 &0.503&0.959&0.521&0.564& 0.481  & 0.460& 0.458\\
 &&&(0.062) &(0.144)&(0.068)&(0.067)&(0.045)&(0.007)&({\bf0.000})\\
 \hline
 GK03 & 25  & 150 &0.517&1.002&0.531&0.517& 0.452  & 0.366& 0.374\\
 &&&(0.060) &(0.140)&(0.077)&(0.066)&(0.042)&(0.007)&({\em0.034})\\
 \hline
 GK04 & 50  & 150 &0.712&1.164&0.748&0.706& 0.669  & 0.528& 0.527\\
 &&&(0.090) &(0.143)&(0.098)&(0.079)&(0.081)&(0.021)&({\em0.027})\\
 \hline
 GK05 & 25  & 200 &0.462&1.124&0.552&0.493& 0.397  & 0.294& 0.289\\
 &&&(0.072) &(0.153)&(0.118)&(0.087)&(0.046)&(0.004)&({\em0.012})\\
 \hline
 GK06 & 50  & 200 &0.703& 1.236&0.751&0.714& 0.611  & 0.429& 0.417\\
 &&&(0.070) &(0.141)&(0.108)&(0.077)&(0.060)&(0.018)&({\bf0.015})\\
 \hline
 GK07 & 25  & 500 &0.523&1.468&0.651&0.496& 0.382  & 0.093& 0.111\\
 &&& (0.088)&(0.092)&(0.087)&(0.089)&(0.082)&(0.004)&({\em0.005})\\
 \hline
 GK08 & 50  & 500 &0.749&1.517&0.835&0.749& 0.534  & 0.166& 0.169\\
 &&&(0.086) &(0.109)&(0.125)&(0.085)&(0.066)&(0.006)&({\em0.013})\\
 \hline
 GK09 & 25  & 1500 &0.890&2.312&1.064&0.695& 0.558 & 0.029& 0.030\\
 &&&(0.075) &(0.113)&(0.133)&(0.070)&(0.042) &(0.001)&({\bf0.001})\\
 \hline
 GK10 & 50  & 1500 &1.101&1.883&1.177&0.950& 0.727  & 0.052& 0.053\\
 &&&(0.065) &(0.076)&(0.082)&(0.090)&(0.070)&(0.003)&({\bf0.002})\\
 \hline
 GK11 & 100  & 2500 &1.237&1.677&1.246&1.161& 0.867  & 0.052& 0.056\\
 &&&(0.060) &(0.056)&(0.067)&(0.063)&(0.061)&(0.002)&({\bf0.002})\\
\hline
\multicolumn{3}{c} {average}& 0.605 & 1.597 & 0.631 & 0.595 & 0.493  & 0.329 & 0.331\\
&   &   &  (0.057) & (0.215) & (0.068) & (0.057) &(0.043) &(0.012) & ({\em0.015})\\
\hline
\end{tabular}
\centering \caption{Average gaps of best solutions and their
standard deviations of the RWCEA and other EAs}
\end{table}

Although some commercial integral linear programming (ILP)
solvers, such as \texttt{CPLEX}, can solve ILP problems with
thousands of integer variables or even more, it seems that the
MKP remains rather difficult to handle when an optimal solution
is wanted. To CB-suit, the results in \cite{chu199863} showed
that major instances of this suit cannot be solved in a
reasonable amount of CPU time and memory by \texttt{CPLEX}. To
GK-suit, which includes still more difficult instances with $n$
up to $2500$, Fr\'{e}ville (2004) in \cite{freville20041}
mentioned that \texttt{CPLEX} cannot tackle these instances.
Therefore, it appears that the MKP continues to be a challenging
problem for commercial ILP solvers.

The best known solutions to these benchmarks, as far as we
known, were obtained by Vasquez and Hao
(2001)\cite{vasquez2001328} and was improved by Vasquez and
Vimont (2005)\cite{vasquez200570}. Their method is based on tabu
search and time-consuming compared with EA.

Raidl and Gottlieb (2005)\cite{raidl2005441} tested six different
variants of EAs, which are called Permutation Representation (PE),
Ordinal Representation (OR), Random-Key Representation (RK),
Weight-Biased Representation (WB), i.e. Raidl's WCEA, and Direct
Representation (DI and DIH). We compare the RWCEA with these EAs
except DIH first. We use all GK-suite and draw out nine instances
(called CB1 to CB9) from CB-suite, which are the first instances
with $\alpha=0.5$ for each combination of $m$ and $n$.

For a solution $x$, the {\em gap} is defined as: \[
gap=\frac{f(x^{LP})-f(x)}{f(x^{LP})} \] where $x^{LP}$ is the
optimum of the LP-relaxed problem to measure the quality of $x$.

We implement the RWCEA on a personal computer (Inter
Core$^{\text{TM}}$ Duo T5800, 2 GHz, 1.99 GB main memory, Windows
XP) using \texttt{DEV-C++}. The initial population is generated by
\texttt{MATLAB}. The population size is 100, and each run was
terminated after $10^6$ created solution candidates; rejected
duplicates were not counted.

Table 1 shows the average gaps of the final solutions and their
standard deviations obtained from independent 30 runs per problem
instance obtained by the RWCEA and other six variants. The results
of other six variants come from \cite{raidl2005441}. In the last
column, bold fonts mean that the results of RWCEA is the best (or
equally best) in the seven EAs. Italics in the last column mean that
the results of RWCEA is better or equal than PE, OR, RK, DI, and
WCEA but slightly worse than DIH. From this table we can draw the
conclusion that the RWCEA is an improvement of WCEA. Especially in
GK02 to GK11, the RWCEA performed much better than Raidl's method.


Table 1 also shows that the RWCEA performed averagely slightly worse
than DIH. But we will point out that can yield better results than
DIH in some instances. Since the best results can be obtained by
CPLEX in CB-suite when $\{m,\,n\}=\{5,\,100\}$, $\{10,\,100\}$, and
$\{5,\,250\}$, we tested the other 180 instances in CB-suite. Each
instance was computed 30 times and the best results were compared
with the results reported in OR-library. The data of the numbers
that the RWCEA yielded better, equal or worse results than the
results reported in OR-library is shown in Table 2. Tables 3 to 8
show the comparison of each instance. These tables show that the
results of more than 50\% instances can be improved by the RWCEA.

\begin{table}[!h]
\begin{tabular}[c]{cccccc}
  \hline
  $m$ & $n$ & number of the instance & better & equal & worse\\
  \hline
  30 & 100 & 30 & 2 & 28 & 0\\
  10 & 250 & 30 & 12 & 16 & 2\\
  30 & 250 & 30 & 15 & 10 & 5\\
  5  & 500 & 30 & 19 & 9 & 2\\
  10 & 500 & 30 & 23 & 4 & 3\\
  30 & 500 & 30 & 21 & 4 & 5\\
  \hline
  \multicolumn{2}{c}{Total} & 180 & 92 & 71 & 17\\
  \hline
\end{tabular}
\centering \caption{The data of the numbers that the RWCEA yielded
better, equal and worse results than the results reported in
OR-library}
\end{table}

\section{Conclusion} We have proposed a RWCEA for solving
multidimensional knapsack problems. This RWCEA has been different
from Raidl's WCEA in the ways that surrogate multipliers are not
used and a heuristic method is incorporated in initialization.
Experimental comparison has shown that the RWCEA can yield better
results than Raidl's WCEA in \cite{raidl1999596} and better results
than the ones reported in the OR-library to some existing
benchmarks. So we think this RWCEA is a good opinion in solving
MKPs. A more detailed investigation of the working mechanism of the
RWCEA and the application of RWCEA to other variants of knapsack
problems (such as multiple choice multidimensional knapsack
problems) will be the subjects of further work.

\newpage
\begin{table}
\begin{tabular}[c]{cccccc}
\hline
CB & $\text{OR}_{\text{CB}}$ & RWCEA & CB & $\text{OR}_{\text{CB}}$ & RWCEA\\
\hline
30.100.00  &   21946 & 21946 & 30.100.15  &   41058 & 41058 \\
30.100.01  &   21716 & 21716& 30.100.16  &   41062 & 41062\\
30.100.02  &   20754 & 20754 & 30.100.17  &   42719 & 42719 \\
30.100.03  &   21464 & 21464 & 30.100.18  &   42230 & 42230 \\
30.100.04  &   21814 & 21814 & 30.100.19  &   41700 & 41700 \\
30.100.05  &   22176 & 22716 & 30.100.20  &   57494 & 57494\\
30.100.06  &   21799 & 21799 & 30.100.21  &   60027 & 60027\\
30.100.07  &   21397 & 21397 & 30.100.22  &   58025 & 58025\\
30.100.08  &   22493 & 22493 & 30.100.23  &   60776 & 60776\\
30.100.09  &   20983 & 20983 & 30.100.24  &   58884 & 58884\\
30.100.10  &   40767 & 40767 & 30.100.25  &   60011 & 60011\\
30.100.11  &   41304 & 41304 & 30.100.26  &   58132 & 58132\\
30.100.12  &   41560 & {\bf41587} & 30.100.27  &   59064 & 59064\\
30.100.13  &   41041 & 41041 & 30.100.28  &   58975 & 58975\\
30.100.14  &   40872 & {\bf40889} & 30.100.29  &   60603 & 60603 \\
\hline
\end{tabular}
\centering \caption{The results of CB-suite reported in OR-library
($\text{OR}_{\text{CB}}$) and the ones obtained by the RWCEA
($m=30$, $n=100$)}
\end{table}

\begin{table}
\begin{tabular}[c]{cccccc}
\hline
CB & $\text{OR}_{\text{CB}}$ & RWCEA & CB & $\text{OR}_{\text{CB}}$ & RWCEA\\
\hline
10.250.00  &   59187 & 59187 & 10.250.15  &   110841 & 110841 \\
10.250.01  &   58662 & {\bf58708}& 10.250.16  &   106075 & 106075\\
10.250.02  &   58094 & 58094 & 10.250.17  &   106686 & 106686 \\
10.250.03  &   61000 & 61000 & 10.250.18  &   109825 & 109825 \\
10.250.04  &   58092 & 58092 & 10.250.19  &   106723 & 106723 \\
10.250.05  &   58803 & 58803 & 10.250.20  &   151790 & {\bf151801}\\
10.250.06  &   58607 & {\bf58704} & 10.250.21  &   147822 & 148772\\
10.250.07  &   58917 & {\bf58930} & 10.250.22  &   151900 & 151900\\
10.250.08  &   59384 & {\it59382} & 10.250.23  &   151275 & {\bf151281}\\
10.250.09  &   59193 & {\bf59208} & 10.250.24  &   151948 & {\bf151966}\\
10.250.10  &   110863 & {\bf110913} & 10.250.25  &   152109 & 151209\\
10.250.11  &   108659 & {\bf108702} & 10.250.26  &   153131 & 153131\\
10.250.12  &   108932 & 108932 & 10.250.27  &   153520 & {\bf153578}\\
10.250.13  &   110037 & {\it110034} & 10.250.28  &   149155 & {\bf149160}\\
10.250.14  &   108423 & {\bf108485} & 10.250.29  &   149704 & 149704 \\
\hline
\end{tabular}
\centering \caption{The results of CB-suite reported in OR-library
($\text{OR}_{\text{CB}}$) and the ones obtained by the RWCEA
($m=10$, $n=250$)}
\end{table}

\begin{table}
\begin{tabular}[c]{cccccc}
\hline
CB & $\text{OR}_{\text{CB}}$ & RWCEA & CB & $\text{OR}_{\text{CB}}$ & RWCEA\\
\hline
30.250.00  &   56693 & {\bf56747} & 30.250.15  &   107246 & {\it107183} \\
30.250.01  &   58318 & {\bf58520}& 30.250.16  &   106308 & {\it106261}\\
30.250.02  &   56553 & 56553 & 30.250.17  &   103993 & 103993 \\
30.250.03  &   56863 & {\bf56930} & 30.250.18  &   106835 & {\it106800} \\
30.250.04  &   56629 & 56629 & 30.250.19  &   105751 & 105751 \\
30.250.05  &   57119 & {\bf57146} & 30.250.20  &   150083 & {\bf150096}\\
30.250.06  &   56292 & {\it56290} & 30.250.21  &   149907 & 149907\\
30.250.07  &   56403 & {\bf56457} & 30.250.22  &   152993 & {\bf153007}\\
30.250.08  &   57442 & {\it57429} & 30.250.23  &   153169 & {\bf153190}\\
30.250.09  &   56447 & 56447 & 30.250.24  &   150287 & 150287\\
30.250.10  &   107689 & {\bf107737} & 30.250.25  &   148544 & 148544\\
30.250.11  &   108338 & {\bf108379} & 30.250.26  &   147471 & 147471\\
30.250.12  &   106385 & {\bf106433} & 30.250.27  &   152841 & {\bf152877}\\
30.250.13  &   106796 & {\bf106806} & 30.250.28  &   149568 & {\bf149570}\\
30.250.14  &   107396 & 107396 & 30.250.29  &   149572 & {\bf149601} \\
\hline
\end{tabular}
\centering \caption{The results of CB-suite reported in OR-library
($\text{OR}_{\text{CB}}$) and the ones obtained by the RWCEA
($m=30$, $n=250$)}
\end{table}

\begin{table}
\begin{tabular}[c]{cccccc}
\hline
CB & $\text{OR}_{\text{CB}}$ & RWCEA & CB & $\text{OR}_{\text{CB}}$ & RWCEA\\
\hline
5.500.00  &   120130 & {\bf120145} & 5.500.15  &   220514 & {\bf220520} \\
5.500.01  &   117837 & {\bf117864}& 5.500.16  &   219987 & {\bf219989}\\
5.500.02  &   121109 & {\bf121118} & 5.500.17  &   218194 & {\bf218215} \\
5.500.03  &   120798 & 120798 & 5.500.18  &   216976 & 216976 \\
5.500.04  &   122319 & 122319 & 5.500.19  &   219693 & {\bf219719} \\
5.500.05  &   122007 & {\bf122009} & 5.500.20  &   295828 & 295828\\
5.500.06  &   119113 & {\bf119127} & 5.500.21  &   308077 & {\bf308083}\\
5.500.07  &   120568 & 120568 & 5.500.22  &   299796 & 299796\\
5.500.08  &   121575 & 121575 & 5.500.23  &   306476 & {\bf306480}\\
5.500.09  &   120699 & {\bf120717} & 5.500.24  &   300342 & 300342\\
5.500.10  &   218422 & {\bf218428} & 5.500.25  &   302560 & {\it302559}\\
5.500.11  &   221191 & {\it221188} & 5.500.26  &   301322 & {\bf301329}\\
5.500.12  &   217534 & {\bf217542} & 5.500.27  &   296437 & {\bf296457}\\
5.500.13  &   223558 & {\bf223560} & 5.500.28  &   306430 & {\bf306454}\\
5.500.14  &   218962 & {\bf218966} & 5.500.29  &   299904 & 299904 \\
\hline
\end{tabular}
\centering \caption{The results of CB-suite reported in OR-library
($\text{OR}_{\text{CB}}$) and the ones obtained by the RWCEA ($m=5$,
$n=500$)}
\end{table}

\begin{table}
\begin{tabular}[c]{cccccc}
\hline
CB & $\text{OR}_{\text{CB}}$ & RWCEA & CB & $\text{OR}_{\text{CB}}$ & RWCEA\\
\hline
10.500.00  &   117726 & {\bf117779} & 10.500.15  &   215013 & {\bf215041} \\
10.500.01  &   119139 & {\bf119181}& 10.500.16  &   217896 & {\bf217911}\\
10.500.02  &   119159 & {\bf119194} & 10.500.17  &   219949 & {\bf219984} \\
10.500.03  &   118802 & {\it118784} & 10.500.18  &   214332 & {\bf214346} \\
10.500.04  &   116434 & {\bf116471} & 10.500.19  &   220833 & {\bf220865} \\
10.500.05  &   119454 & {\bf119461} & 10.500.20  &   304344 & 304344\\
10.500.06  &   119749 & {\bf119777} & 10.500.21  &   302332 & {\bf302333}\\
10.500.07  &   118288 & {\it118277} & 10.500.22  &   302354 & {\bf302408}\\
10.500.08  &   117779 & {\it117750} & 10.500.23  &   300743 & {\bf300747}\\
10.500.09  &   119125 & {\bf119175} & 10.500.24  &   304344 & {\bf304350}\\
10.500.10  &   217318 & 217318 & 10.500.25  &   301730 & {\bf301757}\\
10.500.11  &   219022 & {\bf219033} & 10.500.26  &   304949 & 304949\\
10.500.12  &   217772 & 217772 & 10.500.27  &   296437 & {\bf296457}\\
10.500.13  &   216802 & {\bf216819} & 10.500.28  &   301313 & {\bf301353}\\
10.500.14  &   213809 & {\bf213827} & 10.500.29  &   307014 & {\bf307072} \\
\hline
\end{tabular}
\centering \caption{The results of CB-suite reported in OR-library
($\text{OR}_{\text{CB}}$) and the ones obtained by the RWCEA
($m=10$, $n=500$)}
\end{table}

\begin{table}
\begin{tabular}[c]{cccccc}
\hline
CB & $\text{OR}_{\text{CB}}$ & RWCEA & CB & $\text{OR}_{\text{CB}}$ & RWCEA\\
\hline
30.500.00  &   115868 & {\it115864} & 30.500.15  &   215762 & {\bf215832} \\
30.500.01  &   114667 & {\bf114701}& 30.500.16  &   215772 & {\bf215839}\\
30.500.02  &   116661 & 116661 & 30.500.17  &   216336 & {\bf216419} \\
30.500.03  &   115237 & {\it115228} & 30.500.18  &   217290 & {\bf217302} \\
30.500.04  &   116353 & {\bf116370} & 30.500.19  &   214624 & {\bf214634} \\
30.500.05  &   115604 & {\bf115639} & 30.500.20  &   301627 & {\bf301643}\\
30.500.06  &   113952 & {\bf113983} & 30.500.21  &   299985 & {\it299958}\\
30.500.07  &   114199 & {\bf114230} & 30.500.22  &   304995 & {\bf305062}\\
30.500.08  &   115247 & 115247 & 30.500.23  &   301935 & 301935\\
30.500.09  &   116947 & 116947 & 30.500.24  &   304404 & {\bf304411}\\
30.500.10  &   217995 & {\bf218042} & 30.500.25  &   296894 & {\bf296955}\\
30.500.11  &   214534 & {\bf214557} & 30.500.26  &   303233 & {\bf303262}\\
30.500.12  &   215854 & {\bf215885} & 30.500.27  &   306944 & {\bf306985}\\
30.500.13  &   217836 & {\it217773} & 30.500.28  &   303057 & {\bf303120}\\
30.500.14  &   215566 & {\it215553} & 30.500.29  &   300460 & {\bf300531} \\
\hline
\end{tabular}
\centering \caption{The results of CB-suite reported in OR-library
($\text{OR}_{\text{CB}}$) and the ones obtained by the RWCEA
($m=30$, $n=500$)}
\end{table}
\end{document}